\documentclass[twoside,11pt]{article}

\usepackage{jmlr2e}
\usepackage{amsmath,amssymb,amsfonts}
\usepackage[T1]{fontenc}
\usepackage[latin9]{inputenc}
\usepackage{float}
\usepackage{amsmath}
\usepackage{stmaryrd}
\usepackage{stackrel}
\usepackage{graphicx}
\usepackage{multirow}
\floatstyle{ruled}
\newfloat{algorithm}{tbp}{loa}
\providecommand{\algorithmname}{Algorithm}
\floatname{algorithm}{\protect\algorithmname}

\ShortHeadings{Riemannian Proximal Policy Optimization}{Wang et al.}
\firstpageno{1}

\begin{document}

\title{Riemannian Proximal Policy Optimization}

\author{\name Shijun Wang \email shijun.wang@alibaba-inc.com \\
       \addr Dept. of Artificial Intelligence \\
 		Ant Financial Services Group \\
		Seattle, USA
       \AND
       \name Baocheng Zhu \email baocheng.zbc@antfin.com \\
       \name Chen Li \email na.lc@antfin.com\\
       \name Mingzhe Wu \email mingzhe.wmz@antfin.com \\
       \addr Dept. of Artificial Intelligence \\
 		Ant Financial Services Group \\
		Shanghai, China
		\AND
		\name James Zhang \email james.z@antfin.com\\
		\addr Dept. of Artificial Intelligence \\
 		Ant Financial Services Group \\
		New York, USA
		\AND
		\name Wei Chu \email weichu.cw@alibaba-inc.com \\
		\name Yuan Qi \email yuan.qi@antfin.com \\
		\addr Dept. of Artificial Intelligence \\
 		Ant Financial Services Group \\
		Hangzhou, China
		}

\editor{}

\maketitle

\begin{abstract}
In this paper, We propose a general Riemannian proximal optimization
algorithm with guaranteed convergence to solve Markov decision process
(MDP) problems. To model policy functions in MDP, we employ Gaussian
mixture model (GMM) and formulate it as a non-convex optimization
problem in the Riemannian space of positive semidefinite matrices.
For two given policy functions, we also provide its lower bound on
policy improvement by using bounds derived from the Wasserstein distance
of GMMs. Preliminary experiments show the efficacy of our proposed
Riemannian proximal policy optimization algorithm.
\end{abstract}


\section{Introduction}

Reinforcement learning studies how agents explore/exploit environment,
and take actions to maximize long-term reward. It has broad applications
in robot control and game playing\citep{mnih2015human,silver2016mastering,argall2009survey,silver2017mastering}.
Value iteration and policy gradient methods are mainstream methods
for reinforcement learning \citep{sutton2018reinforcement,li2017deep}. 

Policy gradient methods learn optimal policy directly from past experience
or on the fly. It maximizes expected discounted reward through a parametrized
policy whose parameters are updated using gradient ascent.
Traditional policy gradient methods suffer from three well-known obstacles:
high-variance, sample inefficiency and difficulty in tuning learning
rate. To make the learning algorithm more robust and scalable to large
datasets, Schulman et al. proposed trust region policy optimization
algorithm (TRPO)\citep{schulman2015trust}. TRPO searches for the optimal
policy by maximizing a surrogate function with constraints placed
upon the KL divergence between old and new policy distributions, which
guarantees monotonically improvements. To further improve the data
efficiency and reliable performance, proximal policy optimization
algorithm (PPO) was proposed which utilizes first-order optimization
and clipped probability ratio between the new and old policies \citep{SchulmanWDRK17}.
TRPO was also extended to constrained reinforcement learning. Achiam
et al. proposed constrained policy optimization (CPO) which guarantees
near-constraint satisfaction at each iteration \citep{achiam2017constrained}.

Although TRPO, PPO and CPO have shown promising performance on complex
decision-making problems, such as continuous-control tasks and playing
Atari, as other neural network based models, they face two typical
challenges: the lack of interpretability, and difficult to converge
due to the nature of non-convex optimization in high dimensional parameter
space. For many real applications, data lying in a high dimensional
ambient space usually have a much lower intrinsic dimension. It may
be easier to optimize the policy function in low dimensional manifolds.

In recent years, Many optimization methods are generalized from Euclidean
space to Riemannian space due to manifold structures existed in many
machine learning problems\citep{absil2007trust,absil2009optimization,Vandereycken2013,huang2015broyden,zhang2016riemannian}.
In this paper, we leverage merits of TRPO, PPO, and CPO and propose
a new algorithm called Riemannian proximal policy optimization (RPPO)
by taking manifold learning into account for policy optimization.
In order to estimate the policy, we need a density-estimation function.
Options we have include kernel density estimation, neural networks,
Gaussian mixture model (GMM), etc. In this study we choose GMM due
to its good analytical characteristics, universal representation power
and low computational cost compared with neural networks. It is well-known
that the covariance matrices of GMM lie in a Riemannian manifold of
positive semidefinite matrices.

To be more specific, we model policy functions using GMM first. Secondly,
to optimize GMM and learn the optimal policy functions efficiently,
we formulate it as a non-convex optimization problem in the Riemannian
space. By this way, our method gains advantages in improving both
interpretability and speed of convergence. Please note that Our RPPO
algorithm can be easily extended to any other non-GMM density estimators,
as long as their parameter space is Riemannian. In addition, previously
GMM has been applied to reinforcement learning by embedding GMM in
the Q-learning framework\citep{agostini2010reinforcement}. So it also
suffers from the headache of Q-learning that it can hardly handle
problems with large continuous state-action space.

\section{Preliminaries}

\subsection{Reinforcement learning}

In this study, we consider the following Markov decision process (MDP)
which is defined as a tuple $\left(S,A,P,r,\gamma\right)$, where
$S$ is the set of states, $A$ is the set of actions, $P:S\times A\times S\rightarrow\left[0,1\right]$
is the transition probability function, $r:S\times A\times S\rightarrow \mathcal{R}$
is the reward function, and $\gamma$ is the discount factor which
balances future rewards and immediate ones.

To make optimal decisions for MDP problems, reinforcement learning
was proposed to learn optimal value function or policy. A value function
is an expected, discounted accumulative reward function of a state
or state-action pair by following a policy $\pi$. Here we define
state value function as $v_{\pi}\left(s\right)=E_{\tau\sim\pi}\left[r(\tau)\mid s_{0}=s\right]$
where $\tau=\left(s_{0},a_{0},s_{1},...\right)$ denotes a trajectory
by playing policy $\pi$, $a_{t}\sim\pi\left(a_{t}\mid s_{t}\right)$,
and $s_{t+1}\sim P\left(s_{t+1}\mid s_{t},a_{t}\right)$. Similarly
we define state-action value function as: $q_{\pi}\left(s,a\right)=E_{\tau\sim\pi}\left[r(\tau)\mid s_{0}=s,a_{0}=a\right]$.
We also define advantage function as $A_{\pi}\left(s,a\right)=q_{\pi}\left(s,a\right)-v_{\pi}\left(s\right)$.

In reinforcement learning, we try to find or learn an optimal policy
$\pi$ which maximizes a given performance metric $J\left(\pi\right)$.
Infinite horizon discounted accumulative return is widely used to
evaluate a given policy which is defined as: $J\left(\pi\right)=\underset{\tau\sim\pi}{E}\left[\sum_{t=0}^{\infty}\gamma^{t}r\left(s_{t},a_{t},s_{t+1}\right)\right]$,
where $r\left(s_{t},a_{t},s_{t+1}\right)$ is the reward from $s_{t}$
to $s_{t+1}$ by taking action $a_{t}$. Please note that the expectation
operation is performed over the distribution of trajectories.

In the work of \citep{Kakade02approximatelyoptimal,SchulmanWDRK17},
for two given policies $\pi$ and $\pi^{\prime}$, their expected
accumulative returns over infinite horizon can be linked by the advantage
functions: $J\left(\pi^{\prime}\right)=J\left(\pi\right)+\underset{\tau\sim\pi^{\prime}}{E}\left[\stackrel[t=0]{\infty}{\sum}\gamma^{t}A_{\pi}\left(s_{t},a_{t}\right)\right]$.
By introducing discounted visit frequencies $\rho_{\pi}(s)=P(s_{0}=s)+\gamma P(s_{1}=s)+\gamma^{2}P(s_{2}=s)+...$
\citep{schulman2015trust,achiam2017constrained}, where $s_{0}\sim\rho_{0}$
, $\rho_{0}\colon S\rightarrow \mathcal{R}$ is distribution of initial state
$s_{0}$, we have $J\left(\pi^{\prime}\right)=J\left(\pi\right)+\underset{s}{\Sigma}\rho_{\pi^{\prime}}(s)\underset{a}{\Sigma}\pi^{\prime}(a\mid s)A_{\pi}\left(s_{t},a_{t}\right)$.
To reduce the complexity of searching for a new policy $\pi^{\prime}$
which increases $J\left(\pi^{\prime}\right)$, we replace discounted
visit frequencies $\rho_{\pi^{\prime}}(s)$ to be optimized with old
discounted visit frequencies $\rho_{\pi}(s)$:$L\left(\pi^{\prime}\right)=L\left(\pi\right)+\underset{s}{\Sigma}\rho_{\pi}(s)\underset{a}{\Sigma}\pi^{\prime}(a\mid s)A_{\pi}\left(s_{t},a_{t}\right)$. 

Assume that the policy functions $\pi(a\mid s)$ are parametrized
by a vector $\theta$, $\pi(a\mid s)=\pi_{\theta}(a\mid s)$ . Searching
for new policy $\pi^{\prime}$ is equivalent to searching new parameters
$\theta^{\prime}$ in the parameter space. So we have $L\left(\pi^{\prime}\right)=L\left(\theta^{\prime}\right)=L\left(\theta\right)+\underset{s}{\Sigma}\rho_{\pi_{\theta}}(s)\underset{a}{\Sigma}\pi_{\theta^{\prime}}(a\mid s)A_{\pi_{\theta}}\left(s,a\right)$. 

\subsection{Riemannian space}

Here we give a brief introduction to Riemannian space, for more details see \citep{eisenhart2016riemannian}.Let $M$ be a connected and finite dimensional manifold with dimensionality
of $m$. We denote by $T_{p}M$ the tangent space of $M$ at $p$.
Let $M$ be endowed with a Riemannian metric $\left\langle .,.\right\rangle $,
with corresponding norm denoted by $\parallel.\parallel$, so that
$M$ is now a Riemannian manifold \citep{eisenhart2016riemannian}.
We use $l\left(\gamma\right)=\int_{a}^{b}\bigparallel\gamma^{\prime}\left(t\right)\bigparallel dt$
to denote length of a piecewise smooth curve $\gamma:\left[a,b\right]\longrightarrow M$
joining $\theta^{\prime}$ to $\theta$, i.e., such that $\gamma\left(a\right)=\theta^{\prime}$
and $\gamma\left(b\right)=\theta$. Minimizing this length functional over
the set of all piecewise smooth curves passing $\theta^{\prime}$ and $\theta$, we get a Riemannian distance $d\left(\theta^{\prime},\theta\right)$ which induces
original topology on $M$. Take $\theta\in M,$ the exponential map $exp_{\theta}:T_{\theta}M\longrightarrow M$
is defined by $exp_{\theta}v=\gamma_{v}\left(1,\theta\right)$ which maps a
tangent vector $v$ at $\theta$ to $M$ along the curve $\gamma$. For
any $\theta^{\prime}\in M$ we define the exponential inverse map $exp_{\theta^{\prime}}^{-1}:M\longrightarrow T_{\theta^{\prime}}M$
which is $C^{\infty}$ and maps a point $\theta^{'}$on $M$ to a tangent
vector at $\theta$ with $d\left(\theta^{\prime},\theta\right)=\parallel exp_{\theta^{\prime}}^{-1}\theta\parallel$.
We assume $\left(M,d\right)$ is a complete metric space, bounded
and all closed subsets of $M$ are compact. For a given convex function
$f:M\rightarrow R$ at $\theta^{\prime}\in M$, a vector $s\in T_{\theta^{\prime}}M$
is called subgradient of $f$ at $\theta^{\prime}\in M$ if $f\left(\theta\right)\geq f\left(\theta^{\prime}\right)+\left\langle s,exp_{\theta^{\prime}}^{-1}\theta\right\rangle$,
for all $\theta\in M$. The set of all subgradients of $f$ at $\theta^{\prime}\in M$
is called subdifferential of $f$ at $\theta^{\prime}\in M$ which is denoted
by $\partial f\left(\theta^{\prime}\right)$. If $M$ is a Hadamard manifold
which is complete, simply connected and has everywhere non-positive
sectional curvature, the subdifferential of $f$ at any point on $M$
is non-empty \citep{ferreira2002proximal}.

\section{Modeling policy function using Gaussian mixture model}

To model policy functions, we employ the Gaussian mixture model which
is a widely used and statistically mature method for clustering and
density estimation. The policy function can be modeled as $\pi(a\mid s)=\Sigma_{i=1}^{K}\alpha(s)\mathcal{N}(a;\mu_{i}(s),S_{i}(s))$,
where $\mathcal{N}$ is a (multivariate) Gaussian distribution with mean $\mu\in R^{d}$
and covariance matrix $S\succ0$, $K$ is number of components in
the mixture model, $\alpha=(\alpha_{1},\alpha_{2},...,\alpha_{K})$
are mixture component weights which sum to 1. In the following, we
drop $s$ in GMM to make it simple and parameters of GMM still depend
on state variable $s$ implicitly.

Let's define $\theta=((\alpha_{1},\mu_{1},S_{1}),(\alpha_{2},\mu_{2},S_{2}),...,(\alpha_{K},\mu_{K},S_{K}))$
which parametrizes the policy function. For a given policy function
$\pi_{\theta}(a\mid s)$, we would like to find a new policy $\pi_{\theta^{\prime}}^{\prime}(a\mid s)$
which has higher performance evaluated by $L\left(\pi^{\prime}\right)=L\left(\pi\right)+\underset{s}{\Sigma}\rho_{\pi}(s)\underset{a}{\Sigma}\pi^{\prime}(a\mid s)A_{\pi}\left(s,a\right)$
within close proximity of the old policy to avoid dramatic policy
updates.

Define $g(\theta^{\prime})=-L(\theta^{\prime})$, and $\varphi(\theta^{\prime})=\frac{\beta}{2}d^{2}(\theta^{\prime},\theta)$
which searches new $\theta^{\prime}$ near the proximity of the old
parameter $\theta$. $d(\theta^{\prime},\theta)$ is a similarity
metric which can be Euclidean distance, KL divergence, J-S divergence
or the 2nd Wasserstein distance\citep{arjovsky2017wasserstein}.

We would like to optimize the following problem with corresponding
constraints from GMMs:

\begin{equation}
\underset{\theta^{\prime}=((\alpha_{1}^{\prime},\mu_{1}^{\prime},S_{1}^{\prime}),(\alpha_{2}^{\prime},\mu_{2}^{\prime},S_{2}^{\prime}),...,(\alpha_{K}^{\prime},\mu_{K}^{\prime},S_{K}^{\prime}))}{\min}f(\theta^{\prime})=g\left(\theta^{\prime}\right)+\varphi(\theta^{\prime})
\end{equation}

\[
=-L\left(\theta\right)-\underset{s}{\Sigma}\rho_{\pi_{\theta}}(s)\underset{a}{\Sigma}\left(\stackrel[i=1]{K}{\sum}\alpha_{i}^{\prime}\mathcal{N}(a;\mu_{i}^{\prime},S_{i}^{\prime})\right)A_{\pi_{\theta}}\left(s,a\right)+\frac{\beta}{2}d^{2}(\theta^{\prime},\theta),
\]

s.t. $\stackrel[i=1]{K}{\Sigma}\alpha_{i}^{\prime}=1$, $S_{i}^{\prime}\succ0$,
$i=1,2,...,K$.

We employ a reparametrization method to make the Gaussian distributions
zero-centered. We augment action variables by 1 and define a new variable
vector as $a=[a,1]^{\top}$ with new covariance matrix $S=\left[\begin{array}{cc}
S+\mu\mu^{\top} & \mu\\
\mu^{\top} & 1
\end{array}\right]$\citep{hosseini2015matrix}. 

In the Optimization Problem (1), there is a simplex constraint $\alpha\in\Delta_{K}$.
To convert it to a unconstrained problem, we first define $\eta_{k}=log(\frac{\alpha_{k}}{\alpha_{K}})$,
for $k=1,2,...,K-1$, and let $\eta_{K}=0$ be a constant \citep{hosseini2015matrix}.
Then we have the following unconstrained optimization problem:

\begin{equation}
\underset{\theta^{\prime}=\{\eta^{\prime}=\{\eta_{i}^{\prime}\}_{i=1}^{K-1},S^{\prime}=\{S_{i}^{\prime}\succ0\}_{i=1}^{K}\}}{\min}f(\theta^{\prime})=g\left(\theta^{\prime}\right)+\varphi(\theta^{\prime})
\end{equation}

\[
=-L(\theta)-\underset{s}{\Sigma}\rho_{\pi_{\theta}}(s)\underset{a}{\Sigma}\left(\stackrel[i=1]{K}{\sum}\frac{exp(\eta_{k}^{\prime})}{\Sigma_{j=1}^{K}exp(\eta_{j}^{\prime})}\mathcal{N}(a;S_{i}^{\prime})\right)A_{\pi_{\theta}}\left(s,a\right)+\frac{\beta}{2}d^{2}(\theta^{\prime},\theta).
\]

\section{Riemannian proximal method for policy optimization}

\subsection{Riemannian proximal method for a class of non-convex problems}

In this section, following the work of \citep{khamaru2018convergence},
we tackle a more general class of functions of the form: $f\left(\theta\right)=g\left(\theta\right)-h\left(\theta\right)+\varphi\left(\theta\right)$.
We assume the following assumptions hold:

$\textbf{Assumption 1}$:

(a) The function $g$ is continuously differentiable and its gradient
vector field is Lipschitz continuous with constant $L\geq0$. (b)
The function $h$ is continuous and convex. (c) The function $\varphi\left(\theta\right)$
is proper, convex and lower semi-continuous. (d) The function $f$
is bounded below over a complete Riemannian manifold $M$ of dimension
$m$. (e) Solution set of $\min$ $f(\theta)$ is non-empty and its optimum
value is denoted as $f^{*}$.

$\textbf{Lemma 1}$. Under Assumption 1, assume $u_{k}$ and $v_{k}$
are subgradients of the convex functions $h$ and $\varphi$, respectively.
We have $\theta_{k+1}=exp_{\theta_{k}}(-\alpha_{k}(\nabla g(\theta_{k})-u_{k}+v_{k+1}))$,
and $f(\theta_{k})-f(\theta_{k+1})\ge\frac{1}{2\alpha_{k}}d^2(\theta_{k},\theta_{k+1})$, where $\alpha_k$ is the step size.

Proof of Lemma 1 can be found in the Appendix.

$\textbf{Theorem 1}$. Under Assumption 1, the following statements
hold for any sequence $\left\{ \theta_{k}\right\} _{k\ge0}$ generated
by Algorithm 1:

(a) Any limit point of the sequence $\left\{ \theta_{k}\right\} _{k\ge0}$
is a critical point, and the sequence of function values $\left\{ f(\theta_{k})\right\} _{k\ge0}$
is strictly decreasing and convergent.

(b) For $k=0,1,2,...,N$, we have $\sum_{k=0}^{N}d^{2}(\theta_{k},\theta_{k+1})\le\frac{2(f(\theta_{0})-f^{*})}{L}$.
In addition, if the function $h$ is $M_{h}$-smooth: $\lVert u_{k+1}-u_{k}\rVert\leq M_{h}\times d(\theta_{k},\theta_{k+1})$
where $M_{h}$ is a constant, then

\begin{equation}
\sum_{k=0}^{N}\frac{1}{(L+M_{h}+\frac{1}{\alpha_{k}})^2}\lVert\nabla f(\theta_{k+1})\rVert_{2}^2\le\frac{2(f(\theta_{0})-f^{*})}{L}
\end{equation}

Proof of Theorem 1 can be found in the Appendix.

\begin{algorithm}
\caption{Riemannian proximal optimization algorithm}

1. Given an initial point $\theta_{0}\in M$ and choose step size $\alpha_{k}\in(0, \frac{1}{L}]$,
$k\in\{0,1,2,...\}$.

2. For $k=0,1,2,...$, do

Choose subgradient $u_{k}\in\partial h\left(\theta_{k}\right)$, where
$\partial h\left(\theta\right)$ denotes subdifferential (or subderivatives)
of the convex function $h$ at the point $\theta$. Update

\begin{equation}
\theta_{k+1}=\underset{\theta\in M}{\min}\left\{ \varphi(\theta)+\frac{1}{2\alpha_{k}}d^{2}(\theta,\theta_{k}-\alpha_{k}\left(\nabla g\left(\theta_{k}\right)-u_{k}\right))\right\} .
\end{equation}
\end{algorithm}

\subsection{Lower bound of policy improvement}

Assume we have two policy functions $\pi^{\text{\ensuremath{\prime}}}(a\mid s)=\Sigma_{i}\alpha_{i}^{\prime}\mathcal{N}(a;S_{i}^{\prime})$
and $\pi(a\mid s)=\Sigma_{i}\alpha_{i}\mathcal{N}(a;S_{i})$ parameterized by
GMMs with parameters $\theta^{\prime}=\{\alpha^{\prime}=\{\alpha_{i}^{\prime}\}_{i=1}^{K},S^{\prime}=\{S_{i}^{\prime}\succ0\}_{i=1}^{K}\}$
and $\theta=\{\alpha=\{\alpha_{i}\}_{i=1}^{K},S=\{S_{i}\succ0\}_{i=1}^{K}\}$,
we would like to bound the performance improvement of $\pi^{\text{\ensuremath{\prime}}}(a\mid s)$
over $\pi(a\mid s)$ under limitation of the proximal operator. 

In this study, we choose the Wasserstein distance to measure discrepancy
between policy functions $\pi^{\text{\ensuremath{\prime}}}(a\mid s)$
and $\pi(a\mid s)$ due to its robustness. For two distributions $\mu_{0}$
and $\mu_{1}$ of dimension $n$, the Wasserstein distance is defined
as $W_{2}(\mu_{0},\mu_{1})=\underset{p\in\Pi(\mu_{0},\mu_{1})}{inf}\int_{R^{n}\times R^{n}}\parallel x-y\parallel^{2}p(x,y)dxdy$
which seeks a joint probability distribution $\Pi$ in $R^{2n}$ whose
marginals along coordinates $x$ and $y$ coincide with $\mu_{0}$
and $\mu_{1}$ , respectively. For two Gaussian distributions $N_{0}(\mu_{0},S_{0})$
and $N_{1}(\mu_{1},S_{1})$, its Wasserstein distance is $W_{2}(N_{0},N_{1})^{2}=\parallel\mu_{0}-\mu_{1}\parallel^{2}+trace(S_{0}+S_{1}-2(S_{0}^{1/2}S_{1}S_{0}^{1/2})^{1/2})$.

First we have the following lemma:

$\textbf{Lemma 2}$. Given two policies parametrized by GMMs $\pi^{\text{\ensuremath{\prime}}}(a\mid s)=\Sigma_{i}\alpha_{i}^{\prime}\mathcal{N}(a;S_{i}^{\prime})$
and $\pi(a\mid s)=\Sigma_{i}\alpha_{i}\mathcal{N}(a;S_{i})$, let $f(\pi^{\prime})=\beta\underset{s}{\Sigma}\rho_{\pi}(s)\underset{a}{\Sigma}\left(\stackrel[i=1]{K}{\sum}\frac{exp(\eta_{i}^{\prime})}{\Sigma_{j=1}^{K}exp(\eta_{j}^{\prime})}\mathcal{N}(a;S_{i}^{\prime})\right)A_{\pi}\left(s,a\right)-D_{W_{2}}^{\pi}(\pi^{\prime},\pi)$,
where $D_{W_{2}}^{\pi}(\pi^{\prime},\pi)=\sum_{s}\rho^{\pi}(s)W_{2}(\pi^{\prime},\pi)$,
$W_{2}$ defines Wasserstein distance between two GMMs. Then exist
$\overset{\sim}{\pi}=\underset{\pi^{\prime}}{\arg\max}$ $f(\pi^{\prime})$,
and $f(\overset{\sim}{\pi})\geq f(\pi)=0$.

Lemma 2 can be simply proved by applying Theorem 1.

To reduce computational complexity, we employ discrete Wasserstein
distance by embedding GMMs to a manifold of probability densities
with Gaussian mixture structure as proposed by \citep{chen2019optimal}.
To be more specific, the discrete Wasserstein distance between two
GMMs $\pi^{\text{\ensuremath{\prime}}}(a\mid s)$ and $\pi(a\mid s)$
is:

\begin{equation}
W_{2}(\pi^{\text{\ensuremath{\prime}}},\pi)^{2}=\sum_{i,j}c^{*}(i,j)W_{2}(\mathcal{N}(a;S_{i}^{\text{\ensuremath{\prime}}}),\mathcal{N}(a;S_{j}))^{2},
\end{equation}

where $c^{*}(i,j)=\underset{c\in\prod(\alpha^{\prime},\alpha)}{\arg\min}\sum_{i,j}c(i,j)W_{2}(\mathcal{N}(a;S_{i}^{\text{\ensuremath{\prime}}}),\mathcal{N}(a;S_{j}))^{2}$.

$\textbf{Lemma 3}$. Given two policies parametrized by GMMs $\pi^{\text{\ensuremath{\prime}}}(a\mid s)=\Sigma_{i}\alpha_{i}^{\prime}\mathcal{N}(a;S_{i}^{\prime})$
and $\pi(a\mid s)=\Sigma_{i}\alpha_{i}\mathcal{N}(a;S_{i})$ , their total variation
distance can be bounded as follows:

\begin{equation}
D_{TV}(\pi^{\text{\ensuremath{\prime}}}(a\mid s),\pi(a\mid s))\leq B_{TV}(\pi^{\prime},\pi)=\sum_{i,j}d^{*}(i,j)B_{TV}(\mathcal{N}(a;S_{i}^{\pi^{\prime}}),\mathcal{N}(a;S_{j}^{\pi})),
\end{equation}

where $B_{TV}(N_{0}(\mu,S_{0}),N_{1}(\mu,S_{1}))=\frac{3}{2}\min\{1,\parallel S_{0}^{-1}S_{1}-I\parallel_{F}\}$
for Gaussian distributions $N_{0}(\mu,S_{0})$ and $\mathcal{N}(\mu,S_{1})$\citep{devroye2018total},
and $d^{*}(i,j)=\underset{d\in\prod(\alpha^{\prime},\alpha)}{\arg\min}\sum_{i,j}d(i,j)B_{TV}(\mathcal{N}(a;S_{i}^{\prime}),\mathcal{N}(a;S_{j}))$.
Please note the bound $B_{TV}(\pi^{\prime},\pi)$ actually is the
Wasserstein distance between the discrete distributions $\alpha^{\prime}=\{\alpha_{1}^{\prime},\alpha_{2,}^{\prime}...,\alpha_{K}^{\prime}\}$
and $\alpha=\{\alpha_{1},\alpha_{2,}...,\alpha_{K}\}$ with pairwise
cost defined by the bound of total variation distance between two
Gaussian distributions $B_{TV}(N_{i}(\mu,S_{i}),N_{j}(\mu,S_{j}))$,
$i,j=1,2,...,K$.

With Lemma 2 and Lemma 3, we have the following theorem:

$\textbf{Theorem 2}$. Given two policy functions $\pi^{\prime}$
and $\pi$ parametrized by GMMs $\pi^{\text{\ensuremath{\prime}}}(a\mid s)=\Sigma_{i}\alpha_{i}^{\prime}\mathcal{N}(a;S_{i}^{\prime})$
and $\pi(a\mid s)=\Sigma_{i}\alpha_{i}\mathcal{N}(a;S_{i})$, assume policy
$\overset{\sim}{\pi}$ is parametrized by $\overset{\sim}{\theta}=\underset{\pi^{\prime}}{\arg\max}$
$f(\pi^{\prime})$ as shown in Lemma 2, then we have the following
bound for any $\pi^{\prime}$ within proximity of $\overset{\sim}{\pi}$:

$J(\pi^{\prime})-J(\pi)\geq -2B_{TV}^{{\pi'}}(\pi',\overset{\sim}{\pi})M^{\overset{\sim}{\pi}}+\frac{1}{\beta}D_{W_2}^{\pi}(\overset{\sim}{\pi},\pi)-\frac{2\gamma\epsilon_{\pi}^{\overset{\sim}{\pi}}}{(1-\gamma)}B_{TV}^{{\pi}}(\overset{\sim}{\pi},\pi)$,

where $\epsilon_{\pi}^{\overset{\sim}{\pi}}=\max_{s}\mid E_{a\sim\overset{\sim}{\pi}}A^{\pi}(s,a)\mid$, $M^{\overset{\sim}{\pi}}=\max_{s,a}|A^{\overset{\sim}{\pi}}(s,a)|$, $B_{TV}^{{\pi'}}(\pi^{\prime},\overset{\sim}{\pi})=\sum_{s}\rho^{{\pi'}}(s)B_{TV}(\pi^{\prime},\overset{\sim}{\pi})$
and $B_{TV}^{\pi}(\overset{\sim}{\pi},\pi)=\sum_{s}\rho^{\pi}(s)B_{TV}(\overset{\sim}{\pi},\pi)$
($B_{TV}(\pi^{\prime},\overset{\sim}{\pi})$ and $B_{TV}(\overset{\sim}{\pi},\pi)$
follow the bound definition $B_{TV}(\pi^{\text{\ensuremath{\prime}}},\pi)$
in Lemma 3), $B_{W_{2}}^{\pi}(\overset{\sim}{\pi},\pi)=\sum_{s}\rho^{\pi}(s)W_{d2}(\overset{\sim}{\pi},\pi)$.

Proofs of Lemma 3 and Theorem 2 are shown in the appendix.

\subsection{Implementation of the Riemannian proximal policy optimization method}

Recall that in the optimization problem (2), we are trying to optimize
the following objective function: $\underset{\theta^{\prime}=\{\eta^{\prime}=\{\eta_{i}^{\prime}\}_{i=1}^{K-1},S^{\prime}=\{S_{i}^{\prime}\succ0\}_{i=1}^{K}\}}{\min}f(\theta^{\prime})=g\left(\theta^{\prime}\right)+\varphi(\theta^{\prime})$.

1) \emph{Riemannian Gradient}

$grad_{\mathbf{S_{i}^{\prime}}}g(\theta^{\prime})=\frac{\partial g(\theta^{\prime})}{\partial S_{i}^{\prime}}=-\stackrel[i=1]{K}{\sum}(\underset{s}{\Sigma}\rho_{\pi_{\theta}}(s)\underset{a}{\Sigma}A_{\pi_{\theta}}\left(s,a\right))\frac{exp(\eta_{i}^{\prime})}{\Sigma_{j=1}^{K}exp(\eta_{j}^{\prime})}\times\frac{\partial \mathcal{N}(a;S_{i}^{\prime})}{\partial S_{i}^{\prime}}$,

$\frac{\partial \mathcal{N}(a;S_{i}^{\prime})}{\partial S_{i}^{\prime}}=\mathcal{N}(a;S_{i}^{\prime})\times\frac{1}{2}\left[-S_{i}^{\prime-1}+S_{i}^{\prime-1}aa^{\top}S_{i}^{\prime-1}\right]$,
$i=1,2,...,K$.

Let $a_{i}=-(\underset{s}{\Sigma}\rho_{\pi_{\theta}}(s)\underset{a}{\Sigma}A_{\pi_{\theta}}\left(s,a\right))\mathcal{N}(a;S_{i}^{\prime})$,
$m=1,2,...,K$,

$grad_{\eta_{m}^{\prime}}g(\theta^{\prime})=\frac{\partial g(\theta^{\prime})}{\partial\eta_{m}^{\prime}}=\frac{a_{m}exp(\eta_{m}^{\prime})}{(\Sigma_{j}exp(\eta_{j}^{\prime})}-\Sigma_{i}\frac{1}{(\Sigma_{j}exp(\eta_{j}^{\prime}))^{2}}\left\{ a_{i}exp(\eta_{i}^{\prime})\times exp(\eta_{m}^{\prime})\right\} $.

For Euclidean distance $d(\theta^{\prime},\theta)=\frac{\beta}{2}\parallel\theta^{\prime}-\theta\parallel_{2}^{2}$,
we have 

$\partial_{S_{i}^{\prime}}\varphi(\theta^{\prime})=\frac{\partial}{\partial S_{i}^{\prime}}(\frac{\beta}{2}\sum_{i=1}^{K}d^{2}(S_{i}^{\prime},S_{i}))=\beta(S_{i}^{\prime}-S_{i})$,
$i=1,2,...,K$. For discrete Wasserstein distance $d=W_{d2}^{2}(\theta^{\prime},\theta)$,
we have 

$\partial_{S_{i}^{\prime}}\varphi(\theta^{\prime})=\frac{\beta}{2}\frac{\partial}{\partial S_{i}^{\prime}}(\sum_{i,j}c^{*}(i,j)trace(S_{i}^{\prime}+S_{j}-2(S_{i}^{\prime1/2}S_{j}S_{i}^{\prime1/2})^{1/2}))$,
where 

$c^{*}(i,j)=\underset{c\in\prod(\alpha^{\prime},\alpha)}{\arg\min}\sum_{i,j}c(i,j)W_{2}(\mathcal{N}(a;S_{i}^{\text{\ensuremath{\prime}}}),\mathcal{N}(a;S_{j}))^{2}$,
$i=1,2,...,K$.

2) \emph{Retraction}

With $S_{i,t}^{\prime}$ and $grad_{S_{i,t}^{\prime}}g(\theta^{\prime})$
shown above at iteration $t$, we would like to calculate $S_{i,t+1}^{\prime}$
using retraction. From \citep{cheng2013riemannian}, for any tangent
vector $\eta\in T_{W}M$, where $W$ is a point in Riemannian space
$M$, its retraction $R_{W}\left(\eta\right):=\underset{X\in M}{\arg\min}\parallel W+\eta-X\parallel_{F}$.
For our case $R_{S_{i,t}^{\prime}}\left(-\alpha_{t}(grad_{S_{i,t}^{\prime}}g(\theta^{\prime})+\partial_{S_{i,t}^{\prime}}\varphi(\theta^{\prime}))\right)=\stackrel[i=1]{n}{\sum}\sigma_{i}q_{i}q_{i}^{\top}$,
where $\sigma_{i}$ and $q_{i}$ are the $i$-th eigenvalues and eigenvector
of matrix $S_{i,t}^{\prime}-\alpha_{t}(grad_{S_{i,t}^{\prime}}g(\theta^{\prime})+\partial_{S_{i,t}^{\prime}}\varphi(\theta^{\prime}))$.

$\eta_{i}$, $i=1,2,...,K-1$ are updated using standard gradient
decent method in the Euclidean space. The calculation and retraction
shown above are repeated until $f(\theta^{\prime})$ converges.

\section{Experimental results}

\subsection{Simulation environments and baseline methods}

We choose TRPO and PPO, which are well-known excelling at continuous-control
tasks, as baseline algorithms. Each algorithm runs on the following
3 environments in OpenAI Gym MuJoCo simulator \citep{todorov2012mujoco}:
InvertedPendulum-v2, Hopper-v2, and Walker2d-v2 with increasing task
complexity regarding size of state and action spaces. For each run,
we compute the average reward for every 50 episodes, and report the
mean reward curve and parameters statistics for comparison. 

\subsection{Preliminary results}

\begin{table}
\caption{Number of parameters of each algorithm on each environment with dimensions
listed.}

\begin{tabular}{|c|c|c|c|c|c|}
\hline 
\multirow{2}{*}{Environments} &
\multicolumn{3}{c|}{Number of parameters} &
Dim. of states &
Dim. of actions\tabularnewline
\cline{2-6} \cline{3-6} \cline{4-6} \cline{5-6} \cline{6-6} 
 & RPPO &
TRPO &
PPO &
 &
\tabularnewline
\hline 
InvertedPendulum-v2  &
104 &
124,026  &
124,026 &
4 &
1\tabularnewline
\hline 
Hopper-v2 &
599 &
5,281,434  &
5,281,434  &
11 &
3\tabularnewline
\hline 
Walker2d-v2  &
599 &
40,404,522  &
40,404,522 &
17 &
6\tabularnewline
\hline 
\end{tabular}
\end{table}

In Fig. 1 we show mean reward (column1) for PPO, RPPO and TRPO algorithms
on three MuJoCo environments, screenshots (column2) and probability
density of GMM (column3) for RPPO on each environment. From the learning
curves, we can see that as the state-action dimension of environment
increases (shown in Table 1), both the convergence speed and the reward improvement slow down. This is because the higher dimension the environment sits, the
more difficult the optimization task is for the algorithm. Correspondingly, in the GMM plot, S, A represent the state and the action dimensions respectively, and the probability density is shown in z axis. In the density plot, we can see that as the environment complexity increases, the density pattern becomes more diverse, and non-diagonal matrix terms also show its importance. The probability density of GMM shows that RPPO learns meaningful structure of policy functions. 

TRPO and PPO are pure neural-network-based models with numerous parameters. This makes the model highly vulnerable to overfitting, poor network architecture design and the hyper-parameters tuning. RPPO achieves better robustness by having much fewer parameters. In Table 1 we compare the number of parameters of each algorithm on each environment. It can be seen that GMM has $10^3\sim 10^5$ order fewer parameters as compared with TRPO and PPO.

\begin{figure}[htbp]
\centerline{\includegraphics[width=0.9\columnwidth]{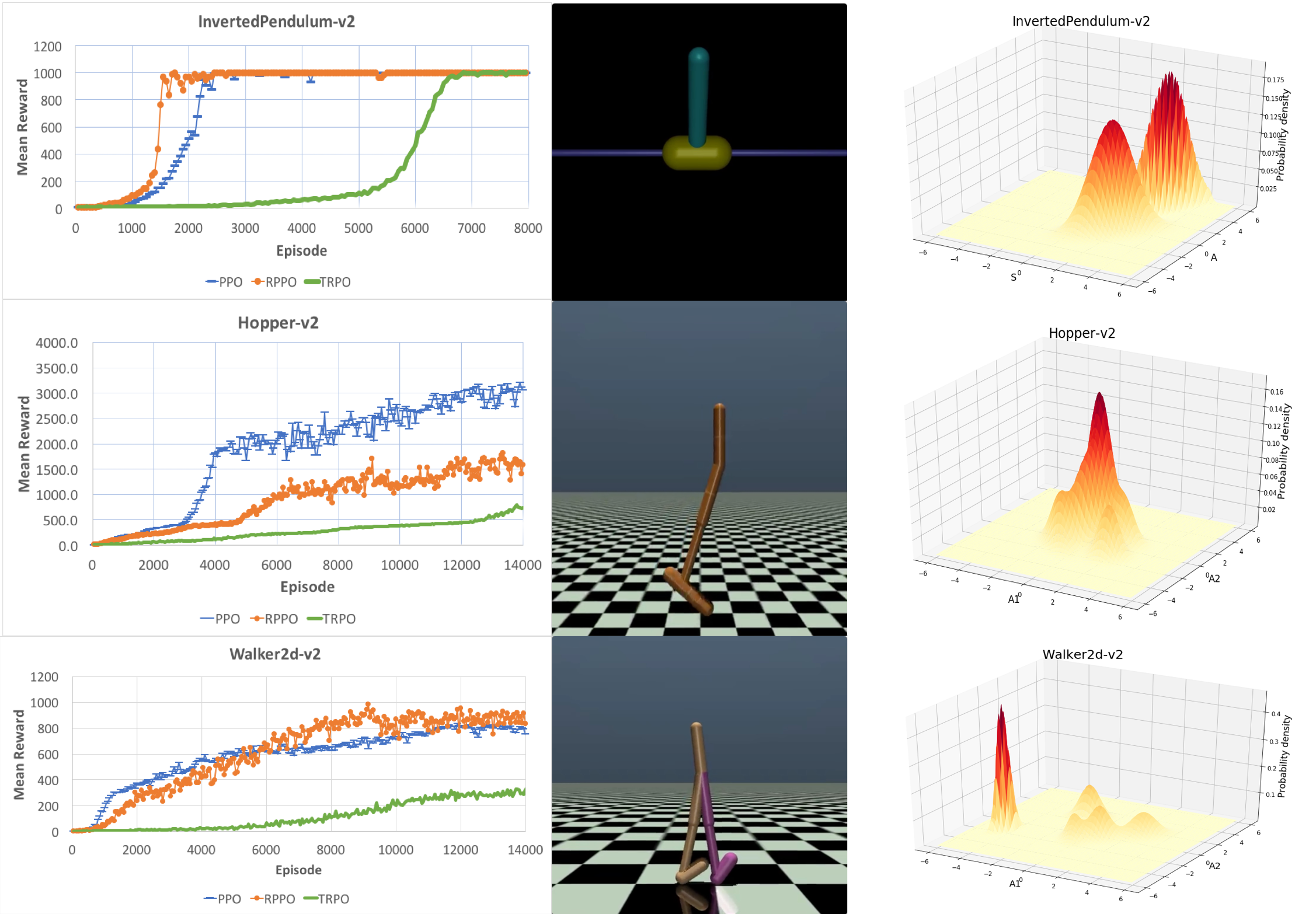}}
\caption{Comparison of PPO, RPPO and TRPO on three MuJoCo environments.}
\end{figure}

\section{Conclusion}

We proposed a general Riemannian proximal optimization algorithm with
guaranteed convergence to solve Markov decision process (MDP) problems.
To model policy functions in MDP, we employed the Gaussian mixture
model (GMM) and formulated it as a non-convex optimization problem
in the Riemannian space of positive semidefinite matrices. Preliminary
experiments on benchmark tasks in OpenAI Gym MuJoCo \citep{todorov2012mujoco}
show the efficacy of the proposed RPPO algorithm.

In Sec. 4.1, the algorithm 1 we proposed is capable of optimizing
a general class of non-convex functions of the form $f\left(\theta\right)=g\left(\theta\right)-h\left(\theta\right)+\varphi\left(\theta\right)$.
Due to page limit, in this study we focus on $f\left(\theta\right)=g\left(\theta\right)+\varphi\left(\theta\right)$
as shown in the Optimization problem (2). In the future, it would
be interesting to incorporate constraints in MDP problems like constrained
policy optimization \citep{achiam2017constrained} and encode them
as a concave function $-h(\theta)$ in our RPPO algorithm.

\appendix
\section*{Appendix}

\subsection*{1. Proof of Lemma 1:}

First let's define a convex majorant $q(w,\theta_{k})$ of the function
$f$ as follows:

$q(w,\theta_{k})=g(\theta_{k})-h(\theta_{k})+\langle \nabla g(\theta_{k})-u_{k},w\rangle +\frac{1}{2\alpha_{k}}\lVert w\lVert_{2}^{2}+\varphi(exp_{\theta_{k}}(w))$,
where $w\in T_{\theta_{k}}M$. Note that minimizer of $q(w,\theta_{k})$ is
the same as $\theta_{k+1}=\underset{\theta\in M}{\min}\left\{ \varphi(\theta)+\frac{1}{2\alpha_{k}}d^{2}(\theta,\theta_{k}-\alpha_{k}\left(\nabla g\left(\theta_{k}\right)-u_{k}\right))\right\} $.
The optimality condition of $\theta_{k+1}$ guarantees that there exists
subgradient $v_{k+1}\in\partial\varphi(\theta_{k+1})$ satisfying the following
equation: $\nabla g(\theta_{k})-u_{k}+v_{k+1}+\frac{1}{\alpha_{k}}w=0$. Let
$w=exp_{\theta_{k}}^{-1}\theta_{k+1}$, we have $\theta_{k+1}=exp_{\theta_{k}}(-\alpha_{k}(\nabla g(\theta_{k})-u_{k}+v_{k+1}))$.

From convexity of the function $\varphi$, for any $\theta\in M$ and $v_{k+1}\in\partial\varphi(\theta_{k+1})$
we have $\varphi(\theta_{k})\geq\varphi(\theta_{k+1})+\langle v_{k+1},w\rangle $, $w\in T_{\theta_{k+1}}M$.
To prove the second inequality in Lemma 1, we have

$f(\theta_{k})-q(w_{k},\theta_{k})=g(\theta_{k})-h(\theta_{k})+\varphi(\theta_{k})-q(w_{k},\theta_{k})$

$\ge g(\theta_{k})-h(\theta_{k})+\varphi(\theta_{k+1})+\langle v_{k+1},w\rangle -q(w_{k},\theta_{k})$

$\ge g(\theta_{k})-h(\theta_{k})+\varphi(\theta_{k+1})+\langle v_{k+1},w\rangle -(g(\theta_{k})-h(\theta_{k})+\langle \nabla g(\theta_{k})-u_{k},w_{k}\rangle +\frac{1}{2\alpha_{k}}\lVert w_{k}\lVert_{2}^{2}+\varphi(exp_{\theta_{k}}(w_{k})))$

$\ge\varphi(\theta_{k+1})+\langle v_{k+1},w\rangle -(\langle \nabla g(\theta_{k})-u_{k},w_{k}\rangle +\frac{1}{2\alpha_{k}}\lVert w_{k}\lVert_{2}^{2}+\varphi(exp_{\theta_{k}}(w_{k}))).$

Since $w=-w_{k}=\alpha_{k}(\nabla g(\theta_{k})-u_{k}+v_{k+1})$, we have

$f(\theta_{k})-q(w_{k},\theta_{k})\ge\langle -\nabla g(\theta_{k})+u_{k}-v_{k+1},w_{k}\rangle -\frac{1}{2\alpha_{k}}\lVert w_{k}\lVert_{2}^{2}$

$\ge\frac{1}{\alpha_{k}}\langle w_{k},w_{k}\rangle -\frac{1}{2\alpha_{k}}\lVert w_{k}\lVert_{2}^{2}=\frac{1}{2\alpha_{k}}\lVert w_{k}\lVert_{2}^{2}$.

Recall that $q(w_{k},\theta_{k})$ is a majorant for the function $f$,
so

$f(\theta_{k})-f(\theta_{k+1})\ge f(\theta_{k})-q(w_{k},\theta_{k})\ge\frac{1}{2\alpha_{k}}\lVert w_{k}\lVert_{2}^{2}=\frac{1}{2\alpha_{k}}d^{2}(\theta_{k},\theta_{k+1})$.

\subsection*{2. Proof of Theorem 1:}

First we would like to prove the convergence of function value. Since
the sequence $\left\{ f(\theta_{k})\right\} _{k\ge0}$ is bounded below,
if $\theta_{k}=\theta_{k+1}$ for some $k$, the convergence of the sequence
$\left\{ f(\theta_{k})\right\} _{k\ge0}$ is trivial. Let's assume that
$\theta_{k}\ne \theta_{k+1}$ for all $k=0,1,2,...$. Under the above assumption,
Lemma 1 ensures that $f(\theta_{k})>f(\theta_{k+1})$. Consequently, there must
exist some scalar $\bar{f}$ which is the limit of $f(\theta_{k})$, $\underset{k\rightarrow\infty}{lim}f(\theta_{k})=\bar{f}$.

Due to page limit, the proof of stationarity of limit points is omitted. 

Now let's establish the bound. Since $f^{*}=\min f(\theta)$ is finite, by
utilizing Lemma 1, we have

$f(\theta_{0})-f^{*}\ge f(\theta_{0})-f(\theta_{N+1})=\sum_{k=0}^{N}(f(\theta_{k})-f(\theta_{k+1}))\ge\sum_{k=0}^{N}\frac{1}{2\alpha_{k}}d^{2}(\theta_{k},\theta_{k+1})$.

Note that $\alpha_{k}\in(0,\frac{1}{L}]$, $k\in\{0,1,2,...\}$, so
$\sum_{k=0}^{N}d^{2}(\theta_{k},\theta_{k+1})\le\frac{2(f(\theta_{0})-f^{*})}{L}$.

Recall that the function $h$ is $M_{h}$-smooth,

$\lVert\nabla f(\theta_{k+1})\rVert_{2}=\lVert \nabla g(\theta_{k+1})-u_{k+1}+v_{k+1}\rVert_{2}=\lVert \nabla g(\theta_{k+1})-u_{k+1}-(\nabla g(\theta_{k})-u_{k}+\frac{1}{\alpha_{k}}d(\theta_{k},\theta_{k+1}))\rVert_{2}$

$\leq\lVert \nabla g(\theta_{k+1})-\nabla g(\theta_{k})\rVert+\lVert u_{k+1}-u_{k}\rVert+\frac{1}{\alpha_{k}}d(\theta_{k},\theta_{k+1})\le(L+M_{h}+\frac{1}{\alpha_{k}})d(\theta_{k},\theta_{k+1})$

So

$\frac{2(f(\theta_{0})-f^{*})}{L}\ge\sum_{k=0}^{N}d^{2}(\theta_{k},\theta_{k+1})\ge\sum_{k=0}^{N}\frac{1}{(L+M_{h}+\frac{1}{\alpha_{k}})^2}\lVert\nabla f(\theta_{k+1})\rVert_{2}^2$

\subsection*{3. Proof of Lemma 3:}

For two Gaussian distributions $N_{0}(\mu,S_{0})$ and $N_{1}(\mu,S_{1})$
with the same mean, their total variation distance is bounded by \citep{devroye2018total}:

$D_{TV}(N_{0}(\mu,S_{0}),N_{1}(\mu,S_{1}))\leq B_{TV}(N_{0}(\mu,S_{0}),N_{1}(\mu,S_{1}))=\frac{3}{2}\min\{1,\parallel S_{0}^{-1}S_{1}-I\parallel_{F}\}$
. 
By using Wasserstein metric, we have

$D_{TV}(\pi^{\text{\ensuremath{\prime}}}(a\mid s),\pi(a\mid s))=\frac{1}{2}\int\mid\Sigma_{i}\alpha_{i}^{\text{\ensuremath{\prime}}}\mathcal{N}(a;S_{i}^{\prime})-\Sigma_{i}\alpha_{i}\mathcal{N}(a;S_{i})\mid da$,

$\leq\sum_{i,j}d^{*}(i,j)B_{TV}(\mathcal{N}(a;S_{i}^{\prime}),\mathcal{N}(a;S_{j}))$,

where $d^{*}(i,j)=\underset{d\in\prod(\alpha^{\prime},\alpha)}{\arg\min}\sum_{i,j}d(i,j)B_{TV}(\mathcal{N}(a;S_{i}^{\prime}),\mathcal{N}(a;S_{j}))$. 

\subsection*{4. Proof of Theorem 2:}

Our proof follows the idea proposed by Wang et al. \citep{wang2018exponentially}.
From Corollary 1 in \citep{achiam2017constrained}, we have

$J(\pi^{\prime})-J(\pi)\geq L_{\pi}(\pi^{\prime})-\frac{2\gamma\epsilon_{\pi}^{\pi^{\prime}}}{(1-\gamma)}D_{TV}^{{\pi}}(\pi^{\prime},\pi)$,
where $\epsilon_{\pi}^{\pi^{\prime}}=\max_{s}\mid E_{a\sim\pi^{\prime}}A^{\pi}(s,a)\mid$, $D_{TV}^{{\pi}}(\pi^{\prime},\pi) = \underset{s}{\Sigma}\rho_{\pi}(s)D_{TV}(\pi^{\text{\ensuremath{\prime}}}(a\mid s),\pi(a\mid s))$
and 
$L_{\pi}(\pi^{\prime})=\underset{s}{\Sigma}\rho_{\pi}(s)\underset{a}{\Sigma}\pi^{'}(a|s)A_{\pi}\left(s,a\right)$.

Similarly, we also have 

$|J(\pi^{\prime})-J(\pi)| = \underset{s}{\Sigma}\rho_{\pi}(s)|\underset{a}{\Sigma}(\pi'(a|s)-\pi(a|s))A_{\pi}\left(s,a\right)|\leq \underset{s}{\Sigma}\rho_{\pi}(s)\underset{a}{\Sigma}|\pi'(a|s)-\pi(a|s)||A_{\pi}\left(s,a\right)|\leq 2D_{TV}^{{\pi'}}(\pi',\pi)M^{\pi}$
where $M^{\pi}=\max_{s,a}|A^{\pi}(s,a)|$.

$J(\pi^{\prime})-J(\pi)=(J(\pi^{\prime})-J(\overset{\sim}{\pi}))+(J(\overset{\sim}{\pi})-J(\pi))$
$\geq -2D_{TV}^{{\pi'}}(\pi',\overset{\sim}{\pi})M^{\overset{\sim}{\pi}}+L_{\pi}(\overset{\sim}{\pi})-\frac{2\gamma\epsilon_{\pi}^{\overset{\sim}{\pi}}}{(1-\gamma)}D_{TV}^{{\pi}}(\overset{\sim}{\pi},\pi)$

$\geq -2D_{TV}^{{\pi'}}(\pi',\overset{\sim}{\pi})M^{\overset{\sim}{\pi}}+\frac{1}{\beta}D_{W_2}^{\pi}(\overset{\sim}{\pi},\pi)-\frac{2\gamma\epsilon_{\pi}^{\overset{\sim}{\pi}}}{(1-\gamma)}D_{TV}^{{\pi}}(\overset{\sim}{\pi},\pi)$.

\vskip 0.2in
\bibliography{rl_gmm_alt}

\end{document}